\documentclass{article}
\usepackage[sglblindworkshop,final,nonatbib]{neurips_2026}
\workshoptitle{ML for Systems 2026}

\usepackage[utf8]{inputenc}
\usepackage[T1]{fontenc}
\usepackage{hyperref}
\usepackage{url}
\usepackage{booktabs}
\usepackage{array}
\usepackage{amsmath,amssymb}
\usepackage{nicefrac}
\usepackage{microtype}
\usepackage{xcolor}
\usepackage{graphicx}
\usepackage{enumitem}
\usepackage{titlesec}
\usepackage{tikz}
\usetikzlibrary{arrows.meta,positioning}
\titlespacing*{\section}{0pt}{5pt}{3pt}
\titlespacing*{\subsection}{0pt}{4pt}{1pt}
\title{WakeKV: Reactive, Reversible KV Residency\\
       for Heads That Change Their Minds}

\author{%
  Utkarsh Ranjan \\
  UC San Diego \\
  \texttt{utranjan@ucsd.edu} \\
}

\begin{document}
\maketitle

\begin{abstract}
Most KV-cache compression methods classify attention heads once, either
offline or during prefill, and keep this classification fixed
throughout generation. Across three models (1.5B--8B) and three
regimes (needle retrieval, long chain-of-thought, and multi-turn
recall), we measure head behavior on four model-regime combinations
and find that most heads change their reading behavior at least once
during generation. We introduce \emph{WakeKV}, a reactive
residency policy that moves cooling heads to a recoverable CPU
reservoir rather than freezing or permanently evicting their state. At
matched memory or budget, WakeKV consistently improves miss rate over
frozen classification and destructive eviction---evaluated across five
model-regime combinations---and over three cited baselines (SnapKV,
uniform R-KV, and ReasonAlloc) across four eligible combinations. A
FlexiCache/vLLM implementation on Mistral-7B confirms the benefit on
real hardware, improving throughput while retaining LongBench quality.
\end{abstract}

\section{Introduction}
\label{sec:intro}

KV-cache size dominates long-context LLM serving memory: caching every
attended key/value pair for every head grows linearly in sequence
length and quickly exceeds GPU memory at scale, so a serving system
must decide which parts of the cache to keep resident, compress, or
evict. The dominant response decides each head's treatment once,
offline or at prefill, and holds that decision fixed for the rest of
generation.

Existing approaches split along two axes: \emph{when} a head's role is
decided, and \emph{whether} a demoted head's state can be recovered.
Static head-role methods (DuoAttention~\cite{duoattention},
RazorAttention~\cite{razorattention}, MoA~\cite{moa},
HeadKV~\cite{headkv}, FlexiCache~\cite{flexicache},
HeteroCache~\cite{heterocache}) and adaptive allocators
(Ada-KV~\cite{adakv}, CAKE~\cite{cake}, LAVa~\cite{lava},
DynamicKV~\cite{dynamickv}) both fix treatment offline or at prefill;
the adaptive allocators do so despite being framed as dynamic.
Decode-time compressors (R-KV~\cite{rkv},
G-KV~\cite{gkv}, ThinKV~\cite{thinkv}, SCOPE~\cite{scope}) update during
generation but evict destructively; ReasonAlloc~\cite{reasonalloc}
reallocates per-head budgets online but still destroys KV on demotion.
Offload systems (ArkVale~\cite{arkvale}, ShadowKV~\cite{shadowkv},
InfiniGen~\cite{infinigen}, HeadInfer~\cite{headinfer}, FlexiCache,
HeteroCache) keep demoted KV recoverable but never revisit head
priority at runtime. Among the systems we survey, none is both
\emph{dynamic} and \emph{reversible} at once (Figure~\ref{fig:mechanism}a).

Recent measurement work on retrieval heads \cite{retrievalheadsdynamic}
casts doubt on this shared premise: attention-head importance is
non-stationary even within a single generation, and that work names
dynamic per-head KV retention as an open direction. We close that gap
with \emph{WakeKV} (\S\ref{sec:method}), a reactive residency policy
that never permanently discards a head's state, and validate it across
five real model/regime combinations at matched memory and budget
(\S\ref{sec:results}).

Our contributions include:
\begin{enumerate}[leftmargin=1.4em,itemsep=1pt,topsep=2pt]
\item \textbf{Measurement:} We show that attention-head behavior
changes during generation, across three models and three regimes,
including two that competing frozen-role methods do not evaluate.
\item \textbf{Prediction study:} We find that inexpensive online
signals do not reliably anticipate these changes.
\item \textbf{WakeKV:} We introduce reactive, reversible KV residency
and show consistent gains over frozen classification, destructive
eviction, and three cited baselines, at matched memory or budget.
\item \textbf{System evaluation:} On A30/Mistral-7B via a
FlexiCache/vLLM shim, WakeKV reactive beats stock throughput at every
rerank interval ($1.34$--$2.33\times$) while holding $95$--$99\%$ of
LongBench quality, confirming the simulator-level direction on real
hardware.
\end{enumerate}

\section{Motivation}
\label{sec:motivation}

Existing KV-cache systems often assume that an attention head's role
remains stable during generation. We ask whether heads instead change
behavior over time, and whether those changes can be predicted well
enough to prefetch their KV state; \S\ref{sec:setup} describes the
experimental setup.

\subsection{Head Churn}
We first measure how much the set of positions attended to by each head changes from one decoding step to the next. Let $A_t$ denote the set of heads (or top-$k$ pages) attended to at
decode step $t$; we measure step-to-step stability using the
adjacent-step Jaccard index:
\begin{equation}
J(A_t, A_{t+1}) \;=\; \frac{|A_t \cap A_{t+1}|}{|A_t \cup A_{t+1}|}.
\label{eq:jaccard}
\end{equation}
$J=1$ means the attended set is unchanged between consecutive steps;
$J=0$ means complete turnover. Measured adjacent-step Jaccard is $0.53$
(continuous score, NIAH), within \cite{retrievalheadsdynamic}'s reported
$0.28$--$0.51$ range (full per-regime breakdown in
Table~\ref{tab:churn}, Appendix).

We then identify ``unstable'' heads using FlexiCache's Random-Corrected
Overlap statistic, which accounts for the amount of overlap that could occur
by chance when selecting from a pool of size $N$:
\begin{equation}
\mathrm{RCO}(A,B) \;=\; \max\!\left(0,\; \frac{\nicefrac{|A\cap B|}{K} - \nicefrac{K}{N}}{1 - \nicefrac{K}{N}}\right), \qquad K = \min(|A|,|B|).
\label{eq:rco}
\end{equation}
$\mathrm{RCO}=0$ corresponds to chance-level overlap, while $\mathrm{RCO}=1$ means that the two sets are identical.  A head is flagged unstable when its mean RCO
over a trailing 16-step window falls below $0.5$. The fraction of heads
with $\geq 1$ HeteroCache-style drift event (windowed-median overlap
dropping below $\tau{=}0.5$, baseline reset after each event) ranges
$0.63$--$0.85$ across regimes and holds at 8B ($0.79$ on CoT). These
two numbers describe different things: only $2$--$9\%$ of heads are
\emph{persistently} unstable (mean RCO below threshold throughout),
while $63$--$85\%$ experience \emph{at least one} discrete drift
event---a small chronically-unstable tail alongside a large majority
of occasionally-shifting heads.

\subsection{Wake-up Prediction}

We grade four candidate signals (page-overlap change, windowed-median
drift, entropy trend, needle-mass delta) under a \emph{causal} per-head
z-score, $z_t = (x_t - \mu_{t-w:t}) / \sigma_{t-w:t}$, computed online
from only the trailing $w{=}32$-step window's mean and standard
deviation. The best signal achieves precision $0.08$--$0.20$ at recall
$0.41$--$0.70$ (an ensemble never beats the single best). A
structurally different candidate---whether the just-generated token
predicts a wake burst (Bonferroni-corrected, held out)---reaches F1
$0.53$--$0.62$ on CoT, $4$--$5\times$ every attention-derived signal,
but zero markers survive on NIAH/multi-turn. Wake-ups are also bursty
($z=+9.7$ vs.\ shuffle null), so a coarse trigger only catches 54\% of
them at a 2\%-of-steps rate. Neither cheap proactive route proved
reliable enough to prefetch on in our experiments.

Together, these findings motivate WakeKV (\S\ref{sec:method}): heads
shift too much for a one-time classification, but not predictably
enough to prefetch---so WakeKV reacts to demand instead of anticipating
it.

\section{Method}
\label{sec:method}

\begin{figure}[t]
\centering
\begin{minipage}[t]{0.40\textwidth}
\centering
\scalebox{0.85}{
\begin{tikzpicture}[
  cell/.style={draw, rounded corners, minimum width=2.7cm, minimum height=1.5cm, align=left, font=\tiny, inner sep=4pt, anchor=north west},
  hdr/.style={font=\tiny\bfseries}
]
\node[hdr] at (1.5,2.15) {offline/prefill};
\node[hdr] at (4.5,2.15) {dynamic/runtime};
\node[cell, fill=black!5] (tl) at (0,1.8) {\textbf{Reversible}\\FlexiCache\\HeteroCache};
\node[cell, fill=blue!8] (tr) at (3.0,1.8) {\textbf{Reversible}\\\textbf{WakeKV} (this paper)\\closes the gap};
\node[cell, fill=black!5] (bl) at (0,0.1) {\textbf{Destructive}\\DuoAttention, HeadKV,\\Ada-KV, CAKE, +4 more};
\node[cell, fill=black!5] (br) at (3.0,0.1) {\textbf{Destructive}\\R-KV, SCOPE,\\ThinKV, ReasonAlloc};
\end{tikzpicture}
}\\[2pt]
{\footnotesize (a) Method landscape by the two axes}
\end{minipage}
\hfill
\begin{minipage}[t]{0.44\textwidth}
\centering
\scalebox{0.68}{
\begin{tikzpicture}[
  box/.style={draw, rounded corners, minimum width=3.0cm, minimum height=0.85cm, align=center, font=\small},
  lbl/.style={font=\scriptsize, align=center},
  >=Latex
]
\node[box, fill=blue!8] (gpu) {GPU resident\\(per-head LRU, budget $B$)};
\node[box, fill=black!5, right=3.4cm of gpu] (cpu) {CPU reservoir\\(recoverable)};
\node[lbl, below=0.55cm of gpu, text width=3.0cm] (destroyed) {$\varnothing$ destroyed\\(unrecoverable)};
\draw[->, thick] (gpu.north) to[bend left=22] node[lbl, above] {demote: over\\budget (LRU)} (cpu.north);
\draw[->, thick] (cpu.south) to[bend left=22] node[lbl, below] {fetch: miss,\\wanted again} (gpu.south);
\draw[->, thick, dashed, black!55] (gpu) -- node[lbl, left, black!55, xshift=-2pt] {evict\\(foil)} (destroyed);
\end{tikzpicture}
}\\[2pt]
{\footnotesize (b) WakeKV's reactive residency mechanism}
\end{minipage}
\caption{(a) Prior methods split along two axes---when role is
decided, and whether demotion is reversible---leaving the
dynamic-and-reversible quadrant empty until WakeKV. (b) A demoted
head's state moves to a recoverable CPU reservoir instead of being
destroyed; the \emph{evict} foil (dashed) shares WakeKV's exact LRU
schedule but destroys state instead of reserving it.}
\label{fig:mechanism}
\end{figure}
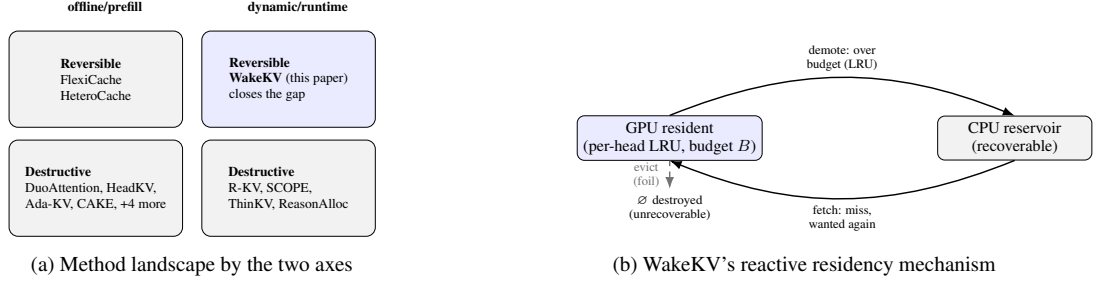

WakeKV replaces per-head classification with a single runtime rule:
each head keeps a GPU-resident set,
LRU-ordered and capped at a per-head budget $B$; an evicted page is
demoted to a CPU reservoir rather than destroyed, and is fetched back
the moment it is wanted again (Figure~\ref{fig:mechanism}b). A wrong or
stale classification thus costs one recoverable stall, not a permanent
accuracy loss.

Concretely, each head maintains a GPU-resident set ordered by
recency, a CPU reservoir holding every demoted page's full,
uncompressed key/value state, and a budget $B$ capping GPU residency.
Each step's attention over resident pages defines \emph{demand}: read
pages move to the front of the LRU order. When demand pushes the
resident set past $B$, the least-recently-used page is
\emph{demoted}---copied to the reservoir, not discarded. A later want
of a non-resident page is a \emph{miss}: its exact state is fetched
back before that step's attention completes, at the cost of one PCIe
transfer, with nothing about its history lost. This single rule both
narrows a head's working set (via $B$) and restores it on demand (via
fetch) without ever committing to a permanent classification.

\section{Evaluation Setup}
\label{sec:setup}

We evaluate on three models, Qwen2.5-3B-Instruct,
DeepSeek-R1-Distill-Qwen-1.5B, and DeepSeek-R1-Distill-Llama-8B, across
three regimes: NIAH (needle retrieval, $\approx$5000-token filler),
long CoT (MATH-500~\cite{math500}, up to 2048 generated tokens), and
multi-turn recall. At every decode step and head we log the top-$k$
attended positions and weights ($k=64$), along with a continuous
needle score and a binary copy-paste score following
\cite{retrievalheadsdynamic}'s Eq.~1--2. The 1.5B and 3B models were
measured on an RTX~2080~Ti (11GB); the 8B model and the real-system
follow-up ran on an A30 (24GB). These logs support the analyses in
\S\ref{sec:motivation} and \S\ref{sec:results} (which adds a fifth
combo, Mistral-7B/NIAH); the real-hardware deployment runs live on a
separate A30.

\section{Results}
\label{sec:results}

Having established that heads shift substantially and cheap
prediction cannot reliably anticipate wake events
(\S\ref{sec:motivation}), we ask whether reacting to demand pays off:
we simulate WakeKV against a frozen classifier and a
destructive-eviction foil sharing its exact LRU schedule, then extend
to three baselines and a real-hardware deployment. Because two
policies at the same nominal budget can realize different memory,
every comparison below holds memory or budget fixed rather than a
shared label. Table~\ref{tab:results} summarizes the headline result;
details follow.

\begin{table}[h]
\centering
\footnotesize
\begin{tabular}{lcc}
\toprule
Comparison (matched memory or budget) & Coverage & WakeKV wins \\
\midrule
vs.\ Frozen, equal memory & 5 model/regime & \textbf{17/17} \\
vs.\ Evict, equal budget & 5 model/regime & \textbf{19/20} \\
vs.\ SnapKV/R-KV/ReasonAlloc, equal memory & 4 model/regime$^\dagger$ & \textbf{32/32} \\
\bottomrule
\end{tabular}
\caption{Headline results: how often reactive's miss rate was $\leq$
the comparator's, at every point where memory or budget was matched
exactly rather than compared at a shared nominal label (full breakdown
in Table~\ref{tab:appendix-sweep}; real-system result in
Table~\ref{tab:m2b2}). $^\dagger$Mistral-7B/NIAH excluded: its
R-KV/ReasonAlloc needed a non-default pruning interval.}
\label{tab:results}
\end{table}

For example, at $\sim$56k matched pages on Mistral/NIAH, reactive's
miss rate is $0.055$ against frozen's $0.405$ ($7\times$ lower); the
one loss against evict is 8B/CoT at $B{=}8$ ($0.632$ vs.\ $0.624$);
and at $\sim$28k matched pages on 8B/CoT, reactive's $0.078$ beats
R-KV's $0.132$.

\subsection{Reactive vs.\ Frozen and Evict}

\begin{figure}[h]
\centering
\includegraphics[width=0.98\textwidth]{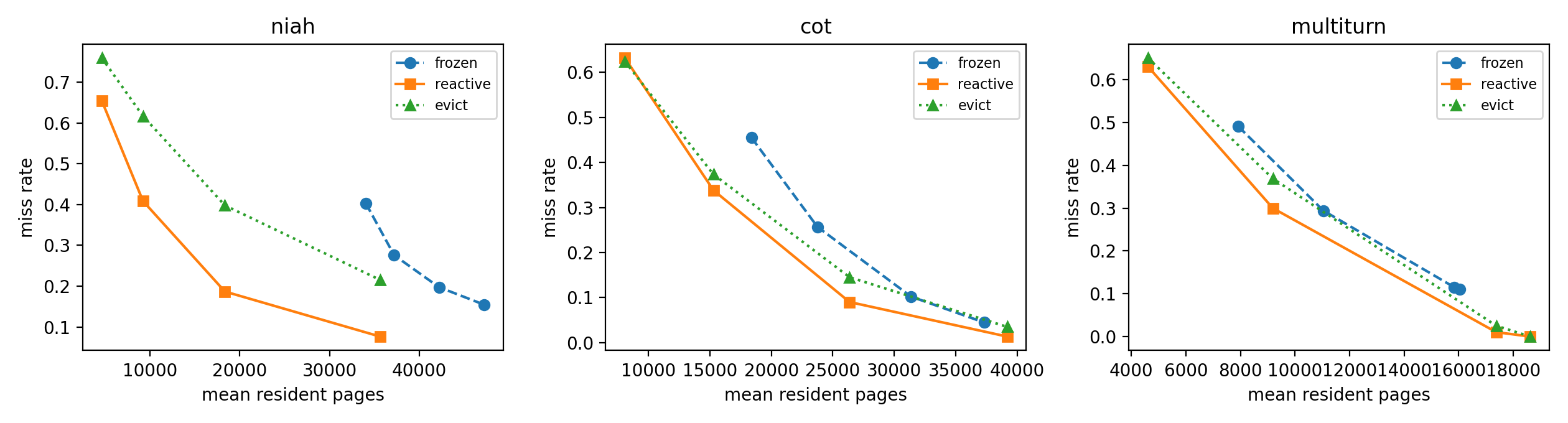}
\caption{Memory-vs-miss-rate Pareto fronts across NIAH, CoT, and
multi-turn. Reactive sits at or below frozen's at all 17
matched-memory points and below evict's at 19 of 20 matched budgets
(Table~\ref{tab:results}).}
\label{fig:pareto}
\end{figure}

We simulate four policies at page granularity over the same logged
attention: \emph{full} (ceiling), \emph{frozen} (FlexiCache's published
defaults), \emph{reactive} (WakeKV, \S\ref{sec:method}), and
\emph{evict} (reactive's exact LRU schedule, but destructive demotion,
isolating reversibility). Sweeping budget $B$ traces each policy's
(memory, miss-rate) curve (Figure~\ref{fig:pareto}); since frozen and
reactive reach different memory at the same $B$, we interpolate
reactive's curve onto frozen's 17 measured memory values (pooled
across combos, dropping 3 of 20 points that reactive never reached).
At those 17
matched-memory levels, reactive's miss rate is $\leq$ frozen's:
\textbf{17/17}. At matched budget (no interpolation needed), reactive
beats evict in \textbf{19/20} settings; the sole exception, at the
tightest 8B/CoT budget, is where recovering a page that evict would
have destroyed alters the subsequent resident set.

We further compare against simulator-compatible reimplementations of
three cited methods---SnapKV~\cite{snapkv}, uniform R-KV~\cite{rkv},
and ReasonAlloc~\cite{reasonalloc} (each a declared simplification;
details in Appendix~\ref{sec:appendix})---at matched rather than
nominal memory, since all three prune periodically and can use several
times reactive's memory at the same $B$
(Table~\ref{tab:appendix-sweep}, ``Reactive mem''). Reactive wins
\textbf{32/32} comparisons across the four eligible combos---e.g.\ on
8B/CoT at $\sim$28k pages of realized memory, R-KV's miss rate is
$0.132$ against reactive's $0.078$ (Table~\ref{tab:results}$^\dagger$).

\subsection{Real-Hardware Deployment}

A FlexiCache/vLLM shim turns stock FlexiCache into WakeKV reactive:
zero heads classified ``stable,'' reranking every step instead of its
native cadence. The strict reactive configuration (rerank every step,
$R{=}1$) is WakeKV's design point, beating stock throughput by
$1.34\times$ at $97.4\%$ LongBench quality. Relaxing how often WakeKV
rechecks head residency trades strictness for throughput, tracing a
frontier that reaches $2.33\times$ at $R{=}16$ while quality stays
above $95\%$ (Table~\ref{tab:m2b2}). Part of the win is less attention
work, not purely memory: zero dense heads at stock's cadence alone
reach that same $2.33\times$.

\begin{table}[h]
\centering
\footnotesize
\begin{tabular}{lrrrr}
\toprule
Config & Rerank int.\ & Gen tok/s & vs stock & LongBench \% \\
\midrule
Stock FlexiCache & 16 (native) & 114.4 & 1.00$\times$ & 100 \\
Reactive (WakeKV) & 1 & 153.6 & \textbf{1.34$\times$} & 97.4 \\
Reactive & 2 & 193.3 & \textbf{1.69$\times$} & 97.1 \\
Reactive & 4 & 226.3 & \textbf{1.98$\times$} & 99.1 \\
Reactive & 8 & 254.2 & \textbf{2.22$\times$} & 97.9 \\
Reactive & $16^\star$ & 266.6 & \textbf{2.33$\times$} & 95.0 \\
\bottomrule
\end{tabular}
\caption{Real-system throughput$\times$quality sweep (A30, Mistral-7B,
LEval 8k$\to$1000; LongBench, 4$\times$30 samples). \textbf{Bold} marks
speedup over stock; $R{=}1$ is WakeKV's design point, $^\star$$R{=}16$
its Pareto knee (quality stays above $95\%$).}
\label{tab:m2b2}
\end{table}

\raggedbottom
\enlargethispage{36pt}
\section{Conclusion and Future Work}

Head importance need not be classified once and frozen. A single
reactive rule---demote under pressure, fetch on demand, never
discard---consistently beats both the frozen and destructive
alternatives (Table~\ref{tab:results}), and the gain carries from
simulation through to real hardware.

Several limitations remain: our evaluation is limited to 1.5B--8B models with
uneven regime coverage; the simulator counts \emph{misses}, not stall
\emph{time}, against FlexiCache-\emph{style} policies rather than
exact code; and the real-system result is narrow (one model/regime,
one A30). A natural next step is comparing reversible and destructive
eviction on a real system using stall time. Still, reactive, reversible
residency is a promising alternative to fixed per-head classification.

\clearpage
\appendix
\par\noindent{\Large\bfseries Appendix}\par\vspace{6pt}
\section{Full Per-Combo, Per-Budget Sweep}
\label{sec:appendix}

Our three baselines are simulator-compatible reimplementations, each a
declared simplification of its cited method: SnapKV~\cite{snapkv}
(prefill-only top-token selection, frozen through decode), uniform
R-KV~\cite{rkv} (periodic redundancy-aware pruning, reduced to
importance-only since raw keys aren't logged), and
ReasonAlloc~\cite{reasonalloc} (online per-head reallocation,
calibrated with an equal split rather than its offline Reasoning-Wave
allocation).

Table~\ref{tab:appendix-sweep} gives the raw miss rate for every policy
at every swept \emph{nominal} budget $B \in \{8,16,32,64\}$, across all
five model/regime combos; this is the source Table~\ref{tab:results}'s
win-counts are computed from. Only Reactive and Evict are directly
comparable within a row, since they share realized memory (identical
LRU cap). Frozen/SnapKV/R-KV/ReasonAlloc realize very different memory
at the same nominal $B$, so Table~\ref{tab:results}'s matched-memory
comparison---not this table---is the appropriate read for those four
columns. Mistral-7B/NIAH's R-KV/ReasonAlloc columns (marked $\dagger$)
additionally use a non-default pruning interval (buffer/delta $=8$,
not the default $128$) to avoid a degenerate no-pruning result on this
model's short NIAH answers, and are excluded from
Table~\ref{tab:results}'s win-counts.

\begin{table}[h]
\centering
\footnotesize
\begin{tabular}{llrrrrrrr}
\toprule
Model/regime & $B$ & Frozen & Reactive & Evict & SnapKV & R-KV & ReasonAlloc & Reactive mem \\
\midrule
Qwen-3B/multiturn & 8  & 0.491 & 0.631 & 0.651 & 0.428 & 0.163 & 0.163 & 4{,}608 \\
                   & 16 & 0.294 & 0.300 & 0.369 & 0.247 & 0.093 & 0.090 & 9{,}187 \\
                   & 32 & 0.116 & 0.010 & 0.025 & 0.201 & 0.010 & 0.009 & 17{,}373 \\
                   & 64 & 0.110 & 0.000 & 0.000 & 0.201 & 0.000 & 0.000 & 18{,}621 \\
\midrule
Qwen-3B/NIAH       & 8  & 0.403 & 0.655 & 0.760 & 0.499 & 0.250 & 0.250 & 4{,}608 \\
                   & 16 & 0.276 & 0.409 & 0.617 & 0.429 & 0.179 & 0.177 & 9{,}209 \\
                   & 32 & 0.197 & 0.187 & 0.398 & 0.374 & 0.128 & 0.126 & 18{,}337 \\
                   & 64 & 0.155 & 0.076 & 0.215 & 0.368 & 0.080 & 0.074 & 35{,}638 \\
\midrule
R1-8B/CoT          & 8  & 0.455 & 0.632 & 0.624 & 0.021 & 0.508 & 0.514 & 8{,}088 \\
                   & 16 & 0.257 & 0.338 & 0.373 & 0.007 & 0.314 & 0.317 & 15{,}298 \\
                   & 32 & 0.102 & 0.090 & 0.145 & 0.005 & 0.132 & 0.129 & 26{,}362 \\
                   & 64 & 0.045 & 0.014 & 0.035 & 0.005 & 0.031 & 0.027 & 39{,}178 \\
\midrule
R1-1.5B/CoT        & 8  & 0.395 & 0.542 & 0.543 & 0.032 & 0.399 & 0.401 & 2{,}637 \\
                   & 16 & 0.191 & 0.221 & 0.263 & 0.014 & 0.201 & 0.199 & 4{,}899 \\
                   & 32 & 0.066 & 0.040 & 0.074 & 0.012 & 0.063 & 0.058 & 8{,}014 \\
                   & 64 & 0.042 & 0.005 & 0.013 & 0.012 & 0.012 & 0.009 & 10{,}692 \\
\midrule
Mistral-7B/NIAH    & 8  & 0.579 & 0.650 & 0.761 & 0.689 & 0.421$^\dagger$ & 0.422$^\dagger$ & 8{,}192 \\
                   & 16 & 0.506 & 0.443 & 0.613 & 0.559 & 0.296$^\dagger$ & 0.303$^\dagger$ & 16{,}349 \\
                   & 32 & 0.442 & 0.180 & 0.308 & 0.442 & 0.167$^\dagger$ & 0.170$^\dagger$ & 32{,}168 \\
                   & 64 & 0.407 & 0.048 & 0.090 & 0.421 & 0.065$^\dagger$ & 0.047$^\dagger$ & 57{,}514 \\
\bottomrule
\end{tabular}
\caption{Raw simulator sweep: miss rate per policy at each nominal
budget $B$ (page units), plus reactive's realized memory (``Reactive
mem''). Reactive vs.\ Evict, the one same-row-comparable pair, favors
reactive in 19 of 20 rows (the exception is 8B/CoT, $B{=}8$); see
\S\ref{sec:results} for why the other four columns require
Table~\ref{tab:results}'s matched-memory correction instead.
$^\dagger$Non-default pruning interval (see text above); excluded from
Table~\ref{tab:results}'s win-counts.}
\label{tab:appendix-sweep}
\end{table}

\clearpage
\section{Head Churn Metrics by Regime}
\label{sec:appendix-churn}

Table~\ref{tab:churn} gives the full per-regime breakdown behind
\S\ref{sec:motivation}'s head-churn numbers: adjacent-step Jaccard, the
fraction of heads that are \emph{persistently} unstable (mean RCO
below threshold throughout), and the fraction that experience
\emph{at least one} HeteroCache-style drift event.

\begin{table}[h]
\centering
\small
\begin{tabular}{lrrrr}
\toprule
Regime & Model & Adj.~Jaccard & Persistently unstable & Ever drifted ($\geq$1 event) \\
\midrule
NIAH & Qwen2.5-3B & 0.53 (cont.) & 0.064 & 0.63 \\
CoT & R1-Distill-1.5B & 0.82 (bin.) & 0.023 & 0.85 \\
Multi-turn & Qwen2.5-3B & 0.79 (bin.) & 0.085 & 0.76 \\
\midrule
CoT & R1-Distill-8B & 0.93 (bin.) & 0.042 & 0.79 \\
\bottomrule
\end{tabular}
\caption{Head-shift metrics by regime. Adjacent Jaccard: continuous
score (NIAH), binary score (CoT/multi-turn). At 8B (bottom row), binary
Jaccard tightens as copy behavior stabilizes on math CoT, but the
page-level drift metric (what a residency controller reacts to)
stays high at 0.79, so the case is undiminished at scale.}
\label{tab:churn}
\end{table}

\end{document}